\documentclass[conference]{IEEEtran}
\IEEEoverridecommandlockouts
\usepackage{cite}
\usepackage{amsmath,amssymb,amsfonts}
\usepackage{optidef}
\usepackage{graphicx}
\usepackage{multirow}
\usepackage{textcomp}
\usepackage{xcolor}
\def\BibTeX{{\rm B\kern-.05em{\sc i\kern-.025em b}\kern-.08em
    T\kern-.1667em\lower.7ex\hbox{E}\kern-.125emX}}
    
\usepackage{mathtools}
\usepackage{wasysym}
\usepackage{bm}
\usepackage{bbm}

\usepackage[font=footnotesize]{subcaption}
\usepackage{booktabs}
\usepackage{array}
\usepackage{dblfloatfix}
\newcolumntype{L}[1]{>{\raggedright\let\newline\\\arraybackslash\hspace{0pt}}m{#1}}
\newcolumntype{C}[1]{>{\centering\let\newline\\\arraybackslash\hspace{0pt}}m{#1}}
\newcolumntype{R}[1]{>{\raggedleft\let\newline\\\arraybackslash\hspace{0pt}}m{#1}}
\newcommand{\subtext}[1]{\text{\textit{#1}}}
\newcommand{\reach}{\subtext{reach}}
\newcommand{\free}{\subtext{free}}
\newcommand{\joints}{\subtext{joints}}
\newcommand{\safe}{\subtext{safe}}
\newcommand{\FK}{\text{\textit{FK}}}
\newcommand{\IK}{\text{\textit{IK}}}
\newcommand{\roll}{\alpha}
\newcommand{\pitch}{\beta}
\newcommand{\yaw}{\gamma}
\newcommand{\link}{\text{\textit{link}}}

\usepackage{balance}
\usepackage{graphicx}

\usepackage{multirow}
\usepackage{tabularx}
\usepackage{enumitem}

\usepackage{algpseudocode}
\usepackage[linesnumbered,ruled,vlined,noend]{algorithm2e}
\SetKwRepeat{Do}{do}{while}

\usepackage{hyperref}
\usepackage{xcolor}
\hypersetup{
    colorlinks,
    linkcolor={red!70!black},
    citecolor={blue!70!black},
    urlcolor={blue!90!black}
}
\usepackage{comment}

\DeclareMathOperator*{\argmin}{arg\,min}

\makeatletter
\let\oldabs\abs
\def\abs{\@ifstar{\oldabs}{\oldabs*}}
\let\oldnorm\norm
\def\norm{\@ifstar{\oldnorm}{\oldnorm*}}
\makeatother

\title{\LARGE \bf
Finding Biomechanically Safe Trajectories for Robot Manipulation of the Human Body in a Search and Rescue Scenario
}
\author{Elizabeth Peiros$^*$, Zih-Yun Chiu$^*$, Yuheng Zhi, Nikhil Shinde, Michael C. Yip \IEEEmembership{Senior Member, IEEE}

\thanks{\footnotesize $^*$ These authors contributed equally to this work. This work was supported by the U.S. Army's Telemedicine and Advanced Technology Research Center under Project W81XWH-22-C-0089.}
\thanks{\footnotesize Elizabeth Peiros, Zih-Yun Chiu, Yuheng Zhi, Nikhil Shinde, and Michael C. Yip are with the Department of Electrical and Computer Engineering, University of California San Diego, La Jolla, CA 92093 USA. {\tt\footnotesize \{epeiros,  zchiu, yzhi, nshinde, m1yip\}@eng.ucsd.edu}}%
}

\begin{document}

\maketitle
\thispagestyle{empty}
\pagestyle{empty}

\begin{abstract}

There has been increasing awareness of the difficulties in reaching and extracting people from mass casualty scenarios, such as those arising from natural disasters. 
While platforms have been designed to consider reaching casualties and even carrying them out of harm's way, the challenge of repositioning a casualty from its found configuration to one suitable for extraction has not been explicitly explored. 
Furthermore, this planning problem needs to incorporate biomechanical safety considerations for the casualty. 
Thus, we present a first solution to biomechanically safe trajectory generation for repositioning limbs of unconscious human casualties. 
We describe biomechanical safety as mathematical constraints, mechanical descriptions of the dynamics for the robot-human coupled system, and the planning and trajectory optimization process that considers this coupled and constrained system.
We finally evaluate our approach over several variations of the problem and demonstrate it on a real robot and human subject.
This work provides a crucial part of search and rescue that can be used in conjunction with past and present works involving robots and vision systems designed for search and rescue.
\end{abstract}


\section{Introduction}


\vspace{-1mm} 
Getting rapid aid and rescue to casualties without putting rescue teams in harm's way is often very challenging or impossible. 
Examples such as the 2023 earthquake in Turkey, with over 43,000 deaths in Turkey and 5,500 in Syria~\cite{engelbrecht_kirac_2023}, present real situations where the scale of medical evacuation is enormous and rapid extraction is critical. Search and rescue robotics promise to offer rescue in scenarios where human medical evacuation may be dangerous and where autonomy is necessary. 

A common scenario involves an unconscious casualty requiring extraction. Mobile manipulation platforms such as the Battlefield Extraction-Assist Robot (BEAR) are designed to extract casualties from disaster and combat zones \cite{theobald2010mobile,watts2004tatrc,murphy2004robot}. However, before they can carry them out, the first task is to reposition the casualty from its found configuration into a feasible configuration for extraction. Inevitably, casualties are found in non-ideal configurations (i.e., poses, such as lying face up, on their side, face down, etc.) that are unsuitable for extraction. This is a current barrier to overcome for extraction robots \cite{williams2019review}. Technical challenges that need to be addressed, of which any one is a deeper research topic, include: 
\begin{enumerate}[leftmargin=*]
\item Parsing visual information to estimate configuration, approximate mass, injuries on the body, vital signs, etc.;
\item Finding optimal grasp candidates on the body based on the visual estimates of kinematics and mass; 
\item Planning the high-level sequence of human body reconfigurations necessary to get from an initial configuration, e.g., crumpled, into an extractable configuration, e.g., supine;
\item Planning, optimizing, and executing an injury-free trajectory to maneuver the human between each intermediate configuration without causing injury.
\end{enumerate}
Problem (1) is a currently active research area in the field of combat medicine~\cite{saputra2019sim,lee2022neural} and fall detection~\cite{alam2022vision}, and (2) can initially be hardcoded with candidate grasp poses on the human body. However, (3) and (4) remain untouched in research literature and relate to solving specifically a \textit{constrained} task-and-motion planning problem~\cite{garrett2021integrated}, well known to be a computationally complex and nuanced problem requiring care in implementation~\cite{qureshi2021constrained}. 
Thus, for this paper, we will address this challenging problem of defining and solving the motion planning and trajectory generation problem for safe human repositioning under biomechanical and geometric constraints. 

\begin{figure}[t]
\vspace{2mm}
\centerline{\includegraphics[width=80mm]{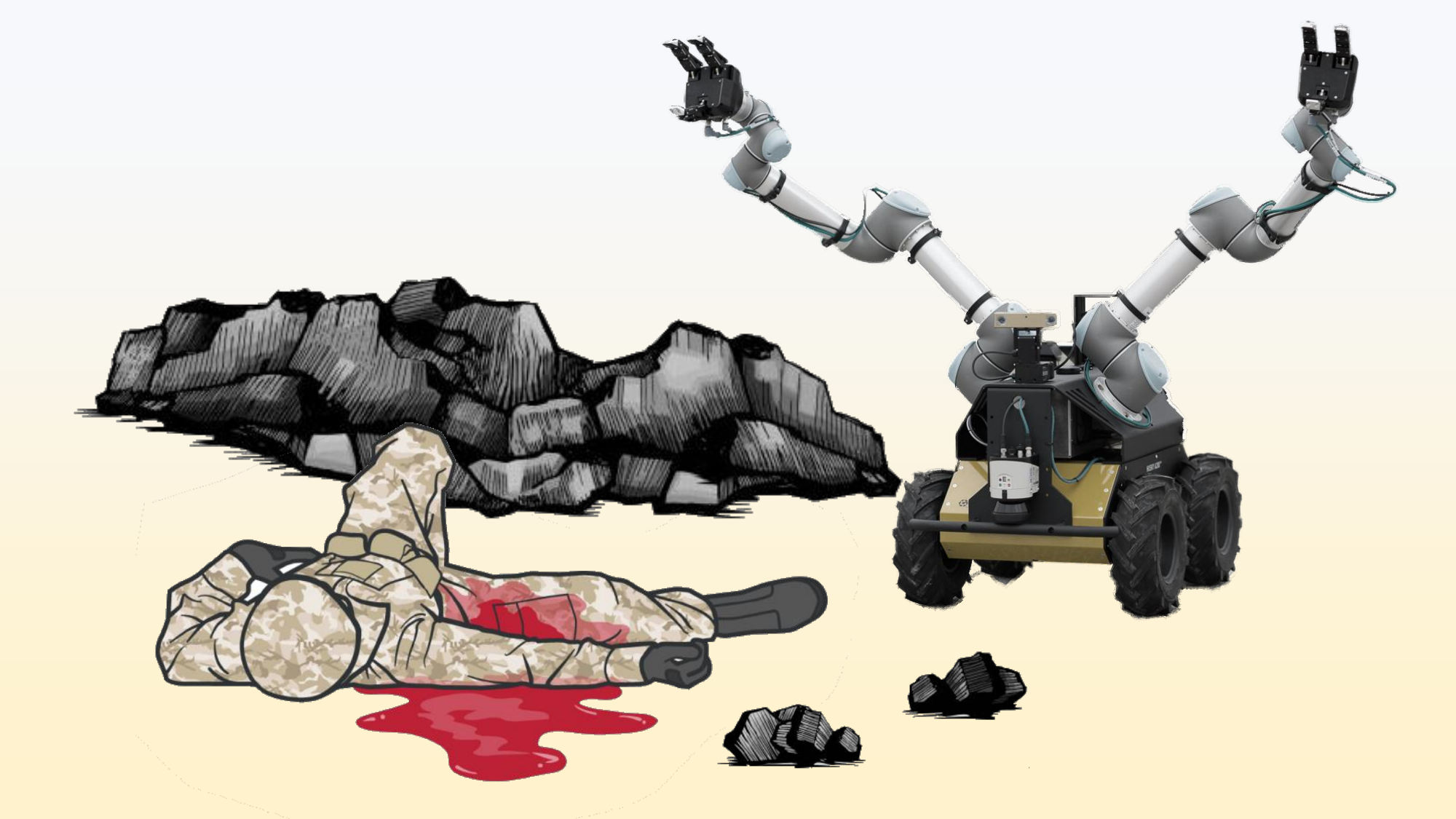}}
\caption{A scenario of finding a casualty with a remote mobile robotic platform. The human is in a non-ideal configuration. This paper provides a first solution towards planning human reconfiguration motions in a biomechanically safe and jointly robot-feasible manner.}
\label{Project_Goal}
\end{figure}

We specifically address the problem of manipulating a passive human in an unanticipated configuration to a desired configuration using a robot manipulator (Fig. \ref{Project_Goal}). We considered both the role of the robot as well as the importance of human safety when the human is incapacitated. Thus, the key contributions of the paper are
\begin{enumerate}[leftmargin=*]
    \item formulation of a constrained trajectory optimization problem for the safe human repositioning problem including a robot as the actuator, 
    \item propose novel musculoskeletal injury avoidance constraints based on biomechanics safety metrics, and 
    \item demonstrate multiple trajectories in simulation and how they achieve safety as well as provide a demonstration of transitioning the solution to a live scenario.
\end{enumerate}

\section{Related works}
The current state-of-the-art in physical human-robot interaction, where the robot is physically manipulating a human, predominantly involves an active/participating human such as in wearables or rehabilitation~\cite{varghese2018wearable,ajoudani2018progress}. In these domains, the focus is much greater on lower-level controllers, such as Impedance and Admittance control, which consider human safety at the cost of scaling back positional accuracy; they also consider the human-in-the-loop as an active member, though for an unconscious person, they would instead be a passive element (acting more as a constraint). To this end, planning problems of contact between robots and sensitive objects involves solving for constrained motion plans~\cite{qureshi2021constrained}.



Biomechanical metrics to define safety have largely resided in physiotherapy applications and have not been formally transferred to safe robotic manipulation. A close example of our problem is an assistive upper extremity device designed to help minimize muscle fatigue for repetitive tasks by providing the user with gravity compensation provided by passive springs~\cite{ZHOU2017337}. Their problem statement minimizes the simulated joint reaction forces while the user moves through a range of motion. Others in the space similarly utilize optimization problem formulations to minimize energy expenditure, often referred to as metabolic cost~\cite{bios11070215,8093939,ong2015simulation}. These devices are aligned with the human anatomy to ensure torque inputs provide motion assistance instead of joint loading. For our problem formulation, we will instead need to address this alignment explicitly given our use of a manipulator.

For devices that involve human interaction, researchers utilize large software packages (Simbody\footnote{https://simtk.org/projects/simbody}, OpenSim\footnote{https://simtk-confluence.stanford.edu:8443/display/OpenSim}, CoBi-Dyn\footnote{https://www.cfd-research.com/products/cobi-dyn/} etc.) that enable the user to gather muscle and joint reaction forces. A collection of these devices and examples of their simulation can be found in~\cite{zhou2020predictive}. These software packages are also excellent for diagnostic muscle-driven human motion planning with injury~\cite{ong2019predicting}. These biomechanics simulation solutions are computationally heavy and can be tricky to converge; thus, motion planning, where forces and torques may need to be re-evaluated thousands of times, would benefit from a reduced-order biomechanics model.

Ultimately, solving the problem of manipulating an unconscious person presents a unique scenario that has not been addressed in prior work. The biggest challenge is that humans cannot actively ensure that their joints are in safe configurations. Thus safety entirely resides on the robot planner and controller to find biomechanically safe motion plans in a reasonable amount of time.

\section{Methods}
\subsection{Problem formulation}
Solving robot motions that move an incapacitated person through biomechanically feasible and safe configurations is a constrained optimization problem. Let $\bm{\theta}\in \mathbb{R}^m$ be the joint angles of a human limb, with $\bm{\theta}_0, \bm{\theta}_T$ the start and goal joint configuration, respectively, and $T$ denotes a time horizon. Let $\bm{q} \in \mathbb{R}^n$ be the joint angles of a robot manipulator with $n$ degrees of freedom. Then the problem is defined as\begin{equation}
\label{eq:florian_opt}
\begin{aligned}
        \textrm{minimize}_{\bm{\theta}_{0:T}} ~~~& J(\bm{\theta}_{0:T})\\
     \text{s.t.} \;\;\; & \bm{\theta}_{0:T} \in \Theta_{\free} \cap \Theta_{\reach} \\
     & \bm{q}_{0:T} \in Q_{\free} \cap Q_{\reach} \\
     & M^H(\bm{\theta}_{0:T}) = 0, \; M^R(\bm{q}_{0:T}) = 0\\
     & \mathbf{C}_g(\bm{\theta}_{0:T}, \bm{q}_{0:T})= 0\\
     & \boldsymbol{f}_H < \boldsymbol{f}_{\safe}
\end{aligned}
\end{equation} 

\noindent where $J$ is the objective to be minimized, $\Theta_{\free}, \Theta_{\reach}$ are the free and reachable human configuration spaces, respectively, $Q_{\free}, Q_{\reach}$ are the free and reachable robot joint spaces, respectively, $M^H(\cdot), M^R(\cdot)$ are the human and robot motion constraints, respectively, $\mathbf{C}_g(\cdot)$ is the grasp constraint, and $\boldsymbol{f}_H$, $\boldsymbol{f}_{\safe}$ are the human-joint reaction forces and maximum allowable force values respectively. 
Note that the reachable human configuration space ensures the human body is moved within its range of motions, and the human motion constraints include constraints that avoid injuring the human body from heavy forces and/or torques being applied from the robot. We assume the robot base pose is fixed in space during motion. 

While the problem formulation in (\ref{eq:florian_opt}) is general for manipulating the entire or any subset of the human body, we focus on moving articulating limbs, and we assume we know a candidate grasp pose (identified apriori for each limb). In this scenario, we will choose an objective to move the grasp pose as minimally as possible as long as it satisfies the constraints of the problem, i.e., 
\begin{multline}
\footnotesize
J(\bm{\theta}_{0:T}) = \sum_{i=1}^T c_p \lVert \bm{p}(\bm{\theta}_i) - \bm{p}(\bm{\theta}_{i-1}) \rVert_2 \\+ c_o \text{Angle}(\bm{o}(\bm{\theta}_i) (\bm{o}(\bm{\theta}_{i-1}))^{-1})
\end{multline}
where $\bm{p}(\bm{\theta})$ and $\bm{o}(\bm{\theta})$ are the positions and axis-angle orientations of the grasp pose in 3D space over the trajectory, and they can be recovered from the forward kinematics of the robot. 
$\text{Angle}(\cdot)$ transforms an axis-angle orientation into an angle difference. 
$c_p$ and $c_o$ are the coefficients for the position and orientation variation, respectively. 
While other objectives may be chosen, this tends to reduce excessive human movement.


\subsection{Robot Motion Constraint}

The robot motion constraint, $M^R(\bm{q}_{0:T})$, is derived from the equations of motion for a serial robotic manipulator under load. This motion constraint is described as
\begin{multline}
    M^R(\bm{q}_i) = D^R(\bm{q}_i)\ddot{\bm{q}_i} + C^R(\bm{q}_i,\dot{\bm{q}}_i) + G^R(\bm{q}_i) \\
    + J^R(\bm{q}_i)^\top \textbf{w}^R_i - \bm{\tau}_i^R, 
\end{multline}
where $D^R(\cdot)$ is the inertia matrix, $C^R(\cdot)$ is the Coriolis term, $G^R(\cdot)$ is the gravitational force, $J^R(\cdot)$ is the robot Jacobian, $\textbf{w}^R_i$ is the generated robot wrench force, and $\bm{\tau}_i$ are the robot joint torques. In path planning, we assume quasi-static, which reduces our equations of motion to:
\begin{equation}
    M^R(\bm{q}_i) = G^R(\bm{q}_i) + J^R(\bm{q}_i)^\top \textbf{w}^R_i - \bm{\tau}_i^R
\end{equation}
In this work, we assume that the robot is much stiffer than humans and has a low-level positional controller running at slow speeds. The robot's differential inertia thus does not apply load on the joints, and we avoid solving robot torques. 

\subsection{Grasping Constraint}

The grasping constraint, $\mathbf{C}_{g}(\cdot)$, transfers the applied forces and torques from the robot to the human.
This transfer can be simply described as a wrench balance: \begin{equation}
    \label{eq:florian_grasp_1}
    \mathbf{C}_{g}(\boldsymbol{\theta}_i,\textbf{q}_i) = \textbf{w}^R_i - \textbf{w}^H_i
\end{equation}
where $\textbf{w}^H_i$ is the wrench force applied on the human at the grasp point. 
Note that $\textbf{w}^R_i$ and $\textbf{w}^H_i$ are defined in the same coordinate frame for the sake of simplicity in equations.

The above balance can be achieved by enforcing the geometric constraint that the end-effector grasp pose of the robot must be equal to the grasp location on the human throughout the entire trajectory, which can be described as
\begin{equation}
    \label{eq:florian_grasp_2}
    \mathbf{C}_{g}(\bm{\theta}_i, \bm{q}_i) = \FK^{H}_{rb}(\bm{\theta}_i) - \FK^{R}_{rb}(\bm{q}_i).
\end{equation}
Here $\FK^{\{H,R\}}_{rb}(\cdot):=\{x,y,z,\roll,\pitch,\yaw\}$ describes the forward kinematics of the human and the robot known apriori based on geometric analysis, and the subscript $rb$ is the robot base frame which is chosen as a common frame of reference. 

\subsection{Human Safety Constraints}

Most biomechanics problems involve using complex and time-consuming models to solve for muscle and joint force/torques. 
These models are unsuitable for search and rescue tasks that are computationally and time-constrained. Instead, we begin with a simplified constraint equation for human dynamics,
\begin{multline}
\label{eq: long human equation}
M^H(\boldsymbol{\theta}_{0:T}) = 
D^H(\boldsymbol{\theta}_i)\ddot{\boldsymbol{\theta}_i} + C^H(\boldsymbol{\theta}_i,\dot{\boldsymbol{\theta}_i}) + G^H(\boldsymbol{\theta}_i) \\+ J^H(\boldsymbol{\theta}_i)^\top\bold{w}_i^{H}  - \bm{\tau}_i^H 
\end{multline}
\noindent where $D^H(\cdot)$ is the human body's inertia, $C^H(\cdot)$ is the Centrifugal/Coriolis term, $G^H(\cdot)$ is the limb gravitational forces, $J^H(\cdot)$ is the human Jacobian, $\textbf{w}^H_i$ is the robot wrench force acting at the grasp pose, and $\bm{\tau}_i^H$ are the human joint torques.

In search and rescue, we note the following assumptions: (1) the unconscious humans cannot produce active force or torques at their joints ($\boldsymbol{\tau}^{H}$ = 0). Joint reaction forces $\boldsymbol{f}^H$ will be calculated in the following section; (2) the reaction forces at the joints will be calculated considering a quasi-static case where neither the body nor robot is moving so quickly as to introduce inertial loads. These assumptions reduce our equation to the gravity forces of the human and the wrench force ($\ddot{\boldsymbol{\theta}} = 0$ and $\dot{\boldsymbol{\theta}} = 0$). The augmented equation appears as 

\begin{equation}
\label{eq: short human equation}
M^H(\boldsymbol{\theta}_{0:T}) = G^H(\boldsymbol{\theta}_i) + J^H(\boldsymbol{\theta}_i)^\top\bold{w}_i^{H}
\end{equation}

\subsubsection{Human Joint Force Modeling}
\label{subsec:human_force_modeling}

To solve for human joint forces, Eq. (\ref{eq: short human equation}) will be used to derive the forces and torques perpendicular to a motion: 
$(\boldsymbol{f}^{\textit{sys}}_{rb})^\top\boldsymbol{u} = 0$ and $(\boldsymbol{m}^{\textit{sys/point}}_{rb})^\top\boldsymbol{u} = 0$. 
$\boldsymbol{f}^{\textit{sys}}_{rb}$ are the forces applied to the system, a singular or set of rigid bodies, and $\boldsymbol{u}$ is a placeholder for the unit vector representing a direction orthogonal to the direction of motion. 
All $\boldsymbol{u}$ should be linearly independent but are not required to be orthogonal to each other. 
The moment equation follows similarly with $\boldsymbol{m}^\textit{sys/point}_{rb}$ representing the moments of the system about a selected point. For each system, these equations can produce up to 6 linearly independent equations.

\begin{figure}[t]
\centerline{\includegraphics[height=45mm]{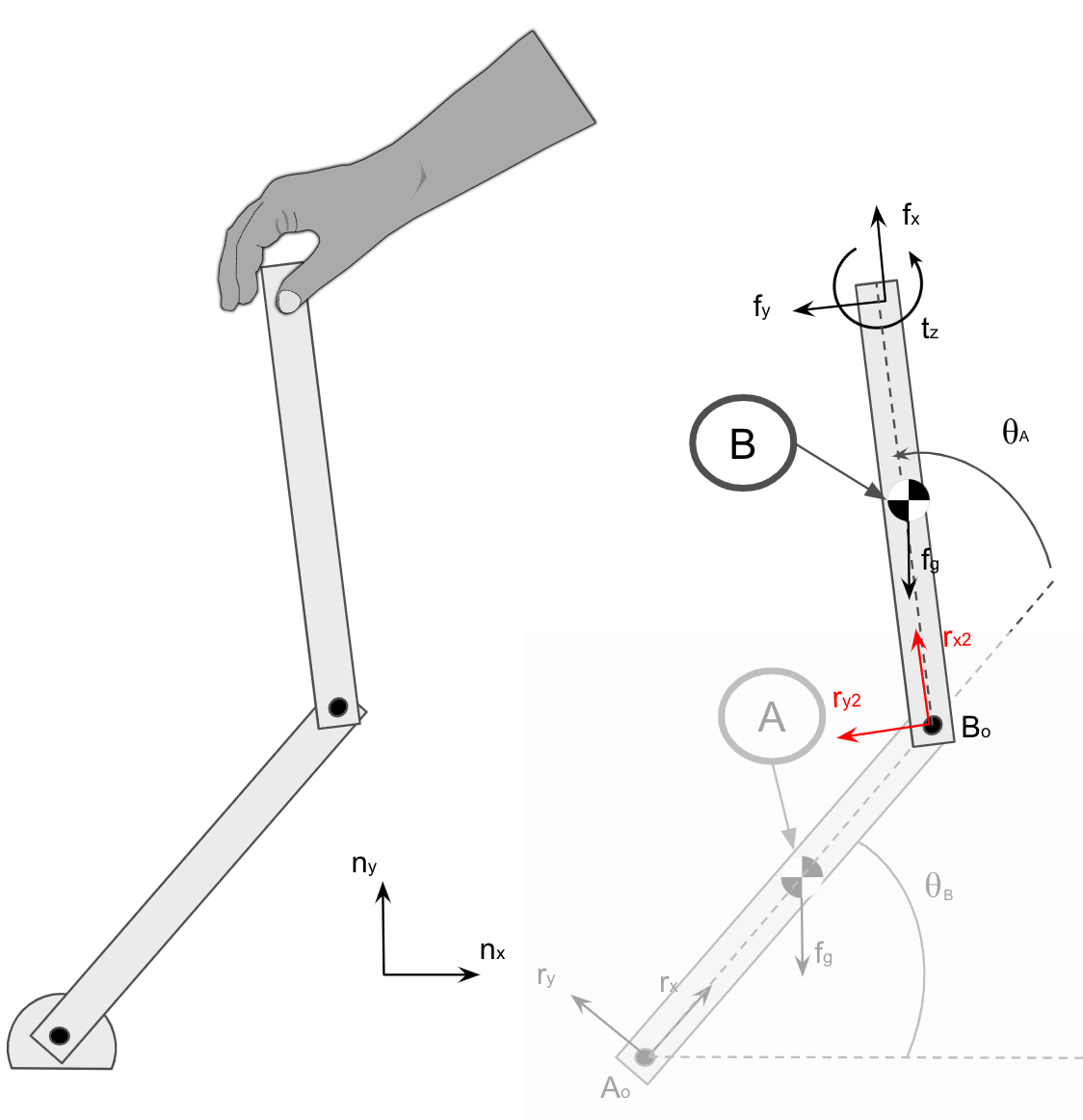} \includegraphics[height=45mm]{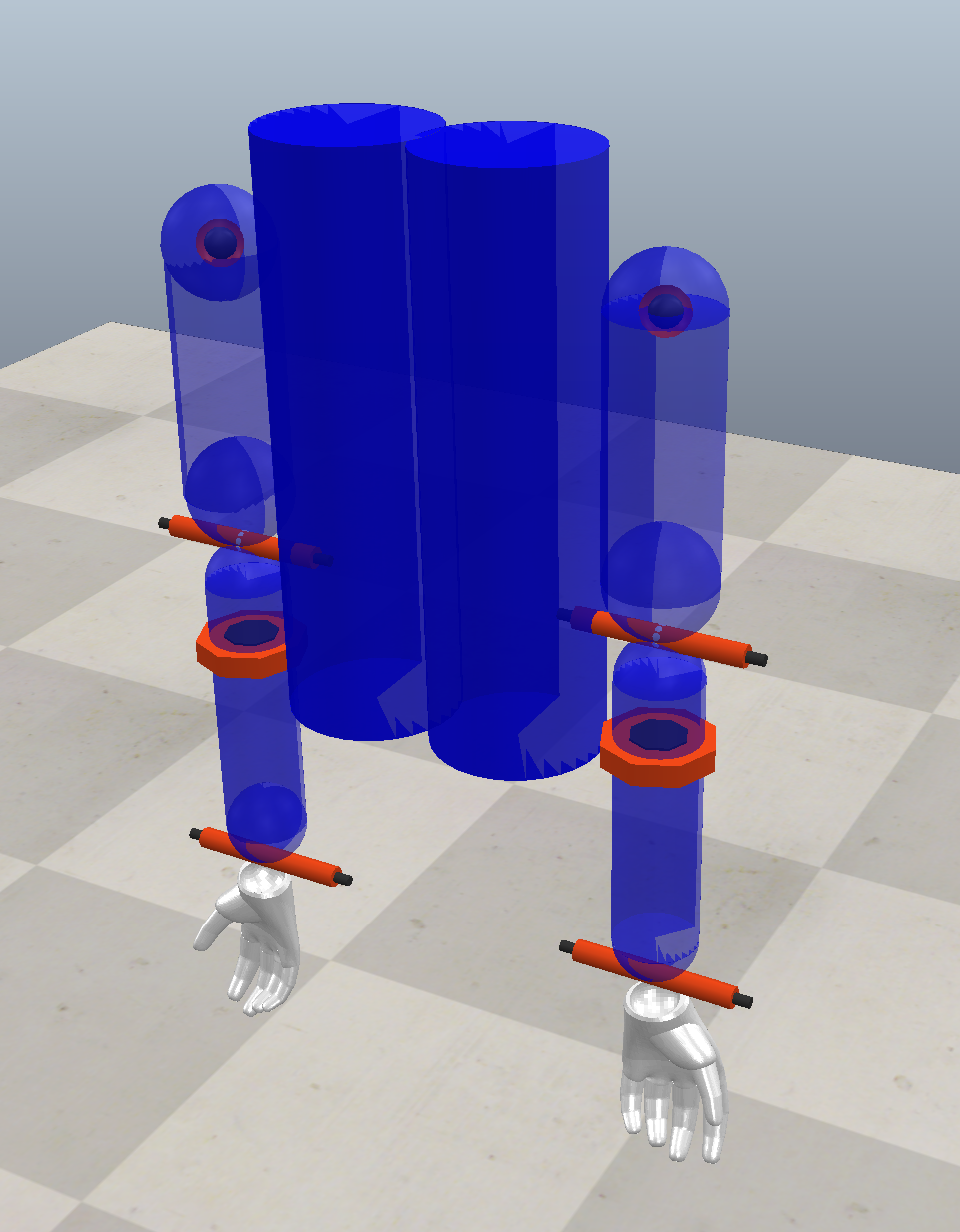} }
\caption{A static model for a 2D Arm is shown on the left as an example of how the reaction forces are applied in the 3D model. The FBD of the final link is highlighted with its internal joint reaction forces of $[r_x, \;r_y]$ shown in red. The wrench force is $[f_x,\;f_y,\;t_z]$. This example is used to exemplify the insufficiency of kinematics to absolve the force safety constraints. Our 3D human geometric model is shown on the right.}
\label{Statics_image}
\end{figure}

To clearly motivate the need for dynamic information beyond kinematic motion planning, we demonstrate our problem with an arm model. 
Then through this example, we show the necessity to appropriately execute force safety constraints on our problem.
Fig. \ref{Statics_image} shows the 3D model rendered in a robotic simulator and a 2D model of how forces and torques were applied to the rigid body system solved with the generalized equations above.

The full system of equations contains more unknown variables than linearly independent equations. A total of 12 equations can be formed while 13 variables must be solved for: the shoulder joint reaction forces $[r_{x1}, r_{y1}, r_{z1}]$, the elbow joint reaction forces and torques $[r_{x2}, r_{y2}, r_{z2}, t_{x2}]$ and the wrench force $[f_x, f_y, f_z, t_x, t_y, t_z]$.
To solve this underdetermined system, one technique would be to define another optimization problem with an objective of minimizing joint forces.
Although valid as a solution, solving one optimization problem inside another optimization problem is time-consuming.
As an alternative, we propose adding an extra model constraint to define an additional linearly independent equation. 
Therefore, the number of equations matches that of unknown variables, and the system of equations becomes uniquely solvable. 

The constraint we will be using is to limit the distribution of the reaction forces parallel to the gravity at the two joints. 
A rudimentary choice would be constraining this reaction-force component at one joint to be 0.
A better model is to balance this component on both joints according to the angle between the gravity and the human limb. 
Hence, we model this reaction-force component on the shoulder, $r_{z1}$, and the elbow, $r_{z2}$, as follows: 
\begin{equation}
\label{eq:3D force model 1}
    r_{z1} = r_{z2}\sin\left(\tfrac{\theta}{2}\right)\text{, where } \theta = \cos^{-1}\left((\boldsymbol{u}_z^{rb})^\top(-\boldsymbol{u}_z^{ua})\right)
\end{equation}
Where $\boldsymbol{u}_z^{rb}$ is the unit vector parallel to the gravity, and $-\boldsymbol{u}_z^{ua}$ is the unit vector along the humeral bone of the arm. 
This equation uses the angle between the arm and the normal of the ground to provide a ratio for which joint holds more of the weight. The same analysis can be done for any other articulating limb of the body.
In this work, we consider the following three different choices of reaction-force constraints: $r_{z1} = 0$, $r_{z2} = 0$, and the balanced one as described in Eq. (\ref{eq:3D force model 1}).  

\subsubsection{Kinematic Insufficiency}
Here we show an example of why kinematic information only is not enough to ensure safe reaction forces on all joints. 
Depicted in the left image of Fig. \ref{Statics_image} is a free-body diagram (FBD) of a single link from a 2-link chain with a pin joint at its base and external forces at the "grasp point" from the robot, i.e., the wrench force. The statics equations derived from this system are:
{\small
\begin{eqnarray}
\label{eq:x_force_1d}
    (\boldsymbol{f}^{\link_b})^\top \boldsymbol{u}^{\link_b}_x = f_x + r_x - m_b g \sin(\theta_A +\theta_B) = 0\\
\label{eq:y_force_1d}
    (\boldsymbol{f}^{\link_b})^\top \boldsymbol{u}^{\link_b}_y = f_y + r_y + m_b g \cos(\theta_A +\theta_B) = 0\\
\label{eq:z_torque_1d}
    (\boldsymbol{m}^{\link_b/B_o})^\top \boldsymbol{u}^{\link_b}_z  = L_b f_y + t_z + \tfrac{1}{2}L_b m_b g \cos(\theta_A +\theta_B) = 0
\end{eqnarray}
}

\noindent Eq. (\ref{eq:x_force_1d}) and (\ref{eq:y_force_1d}) are all the forces applied to $\link_b$ and (\ref{eq:z_torque_1d}) are the moments of $\link_b$ about the pin joint, point $B_o$ shown in Fig. \ref{Statics_image}. In these 3 equations, there are 5 unknown variables: $f_x,\ f_y,\ r_x,\ r_y,\ t_z$. 
Among them, $r_x$ and $f_x$ only appear in Eq. (\ref{eq:x_force_1d}). 
Without a constraint on the external force $f_x$, the reaction forces can be any value within $(-F_{\max}, F_{\max})$, where $F_{\max}$ is the maximum force that can be generated by a robot necessary to follow the position trajectory resulting in a high potential for injury.

\subsubsection{MusculoSkeletal Injury}
\label{subsec: Safety Constraints}
Outside of collecting data on the material properties of tendons, there is still much being explored to understand how tendons take load in vivo due to their non-linear nature~\cite{magnusson2008human}. 

For our path planning problem, we will need to define musculoskeletal injuries quantitatively to ensure that tendons do not exceed the elastic region of material deformation. A low estimate (from injury) for shoulder dislocation is approximately 150N given by biomechanics literature~\cite{clouthier2011effect}. 
The shoulder has no reaction torques as it is modeled as a spherical joint with all rotational degrees of freedom. 

In cases where measured injury thresholds are not readily available, such as the elbow joint, we can use tendon mechanics models with measured mechanical properties to estimate their safety limits.
Eq. (\ref{eq:yield_equation}) shows a typical linear relationship between the yield force ($P_{yp}$), the cross-sectional area ($A_o$), and the stress at the yield point ($\sigma_{yp}$):
\begin{equation}
\label{eq:yield_equation}
    \sigma_{yp} = \tfrac{P_{yp}}{A_{o}}
\end{equation}
The range of $\sigma_{yp}$ has been calculated using data collected in-vitro form the supraspinatus tendon~\cite{nakajima1994histologic} ($\sigma_{yp} = 2.8 \pm 0.7\ {N}/{mm^2}$). 
Combining both solved ($*$) and measured ($\dagger$) thresholds of safety, we can now define the human reaction-force constraints, $\boldsymbol{f}_{H} \;<\; \boldsymbol{f}_{\safe}$ as
\begin{equation} \label{ForceLimits}
\footnotesize
        \begin{bmatrix}
           \|\boldsymbol{f}_{s}\|_2 \\
           \|\boldsymbol{t}_{s}\|_2 \\
           \|\boldsymbol{f}_{e}\|_2 \\
           \|\boldsymbol{t}_{e}\|_2
         \end{bmatrix} <
                  \begin{bmatrix}
           150\;N^\dagger\\
           - \\
           400\;N^* \\
           10\;N/m^*
         \end{bmatrix}
\end{equation}

\subsection{Trajectory Generation}

\begin{algorithm}[t!]
  \vspace{0.5mm}
  \footnotesize
  \SetAlgoLined
  \KwIn{initial and final human-limb configurations, $\bm{\theta}_i$ and $\bm{\theta}_f$; the objective function and the set of constraints defined in (\ref{eq:florian_opt}), $J$ and $\mathcal{C}$; a sampling-based motion planning algorithm, \textit{MP}; a trajectory optimization algorithm, $\textit{TO}$; the distribution of the robot-base pose, $\mathcal{N}(\boldsymbol{\mu}_{rb}, \boldsymbol{\Sigma}_{rb})$; the maximum run time for each sub-process, $T_{\max}$; the inverse kinematic function of the robot, $\IK^R_{rb}$; the forward kinematic function of the human, $\FK^H_{rb}$.}
  \KwOut{a human-limb trajectory, $\bm{\theta}_{0:T}$; a robot-arm trajectory, $\bm{q}_{0:T}$}
  \tcp{Run sampling-based motion planning to solve for a human-limb path}
  $\bm{\theta}_{0:T} = \text{\textit{MP}}(\bm{\theta}_i, \bm{\theta}_f, J, \mathcal{C}, T_{\max})$ \\
  \tcp{Run trajectory optimization to refine the human-limb path}
  $\bm{\theta}_{0:T} = \text{\textit{TO}}(\bm{\theta}_{0:T}, J, \mathcal{C}, T_{\max})$ \\
  \tcp{Run rejection sampling to solve for a robot-base pose}
  \Do{(\ref{eq:florian_grasp_2}) is not satisfied}{
    $\mathbf{p}_{rb} \sim \mathcal{N}(\boldsymbol{\mu}_{rb}, \boldsymbol{\Sigma}_{rb})$ \\
    \tcp{Run inverse kinematics to find the robot-arm configurations}
    $\mathbf{q}_{0:T} = \IK^R_{rb}(\mathbf{p}_{rb}, \FK^H_{rb}(\mathbf{\theta}_{0:T}))$ \\
  }
  \caption{Trajectory Generation of Moving a Human Limb with a Robot Arm}
  \label{alg:trajectory_generation}
\end{algorithm}

With all the constraints defined, we solve (\ref{eq:florian_opt}) in three stages: generate a coarse human motion trajectory, refine the human trajectory, and generate a robot motion trajectory. 
This separation of human and robot trajectory solving treats the biomechanical safety of the human as the priority, followed by finding a feasible robot path, and is key to ensuring that a safe solution is found in a reasonable time. 

A sampling-based motion planning algorithm and trajectory optimization are used to solve for the coarse and refined human motion trajectories, $\bm{\theta}_{0:T}$, while satisfying the human motion constraints, $M^H(\cdot)$, respectively.
A sampling-based motion planner quickly finds a feasible path, even in non-convex scenarios~\cite{lavalle2006planning}. 
A trajectory optimizer takes the non-smooth and suboptimal initial path and performs gradient descent (while checking for constraint satisfaction) to converge quickly to a smoother and shorter path, $\bm{\theta}_{0:T}$.

Since the robot is assumed to regulate a feasible joint trajectory, $\bm{q}_{0:T}$, the robot trajectory can be directly solved for by using the human forward kinematics and the optimized $\bm{\theta}_{0:T}$, and using $\FK^H_{rb}$ to define a geometric grasping constraint (\ref{eq:florian_grasp_2}) for each time step, and then applying inverse kinematics of the robot, $\IK_{rb}^R$. 
The only unknown is the robot base to human transformation, $\bm{p}_{rb}^h \in \mathbb{R}^6$, described in the position and axis-angle spaces. This transformation determines whether an inverse kinematic solution can be found or the trajectory is out-of-reach. In search and rescue missions, this transformation describes how the robot should approach the human, and we use rejection sampling to find a feasible $\bm{p}_{rb}^h$.
A transformation is sampled from a normal distribution with mean $\boldsymbol{\mu}_{rb}$ and covariance $\mathbf{\Sigma}_{rb}$.
The feasibility of a sample is evaluated by checking if (\ref{eq:florian_grasp_2}) is satisfied.
While conducting rejection sampling to solve for a feasible robot base to human transformation, the robot joint trajectory, $\bm{q}_{0:T}$, is solved via inverse kinematics from the grasping constraint (\ref{eq:florian_grasp_2}). 
Algorithm \ref{alg:trajectory_generation} describes the complete trajectory generation process.

\section{Experimental Setup and Results}

We evaluate our proposed human model, biomechanic safety constraints, and trajectory-generation method in both simulation and real-world environments. 
For simulation experiments, we visualize the generated trajectories in CoppeliaSim\footnote{https://www.coppeliarobotics.com/}, a robotic simulator, and evaluate its geometric feasibility. 
For real-world experiments, we perform quantitative and qualitative analyses by letting a robot arm runs its planned trajectories while grasping a real human limb (Fig.~\ref{Experiments}).

\subsection{Experimental Setup}
In the experiments, we use the human kinematic model manually designed in a Unified Robot Description Format (URDF) file, which is visualized in the right image of Fig.~\ref{Statics_image}. 
The robot arm that grasps a human limb is a Franka Emika Panda.
The dynamical equations and biomechanic safety constraints in Eq.~(\ref{eq:florian_opt}) are solved using MotionGenesis\footnote{http://www.motiongenesis.com/}. 
Together with the dynamical constraints, a geometrical collision-checking algorithm FCL~\cite{pan2012fcl} is incorporated into both sampling-based motion planning and trajectory optimization to ensure safe trajectories. 
We choose BIT*~\cite{gammell2015batch} implemented in the Open Motion Planning Library~\cite{sucan2012open} as the sampling-based motion planning algorithm. 
Given an initial human-limb trajectory found by BIT*, SLSQP~\cite{kraft1988software, kraft1994algorithm} implemented in Scipy~\cite{virtanen2020scipy} is used to perform trajectory optimization on it. 
Finally, a sampled robot-base pose and its corresponding robot-arm trajectory are evaluated and solved in CoppeliaSim using its inverse-kinematic solver. 

\begin{table}[t]
\centering
\caption{Human model parameters} 
\label{tab:human_parameters}
\begin{tabular}{cccccc}
    \toprule
     \multirow[c]{2}{*}{Parts} & Sphere & \multicolumn{2}{c}{Cylinder} & \multirow[c]{2}{*}{Weight (kg)} \\
     \cmidrule(lr){2-2}\cmidrule(lr){3-4}
     & Radius (m) & Radius (m) & Length (m) &  \\
    \midrule
    Upper Arm & 0.095 & 0.095 & 0.22 & 5 \\
    Lower Arm & 0.075 & 0.075 & 0.25 & 3 \\
    \bottomrule
    \end{tabular}
\end{table}

Table \ref{tab:human_parameters} shows the parameters we use when defining the human model. 
The upper and lower arms of this model are both composed of two spheres and a cylinder, as pictured in the right image of Fig. \ref{Statics_image}. 
When running trajectory generation, we set $c_p = c_o = 1 $ and the maximum run time of each sub-process as 120 seconds. 
The mean of the robot-base poses $\boldsymbol{\mu}_{rb} = [\bar{x}_{rb}\ \bar{y}_{rb}\ 0\ 0\ 0\ \bar{\gamma}_{rb}]^\top$, where $\bar{x}_{rb}$, $\bar{y}_{rb}$, and $\bar{\gamma}_{rb}$ are manually chosen that makes the robot face the human body. 
The covariance of the robot-base pose is defined as $\boldsymbol{\Sigma}_{rb} = \mathrm{diag}(0.01, 0.0025, 10^{-6}, 10^{-6}, 10^{-6}, 0.07)$, where $\mathrm{diag}(\cdot)$ turns a vector into a diagonal matrix. 

\subsection{Simulation Experiments}
\begin{figure}[t]
\centerline{\includegraphics[width=\linewidth]{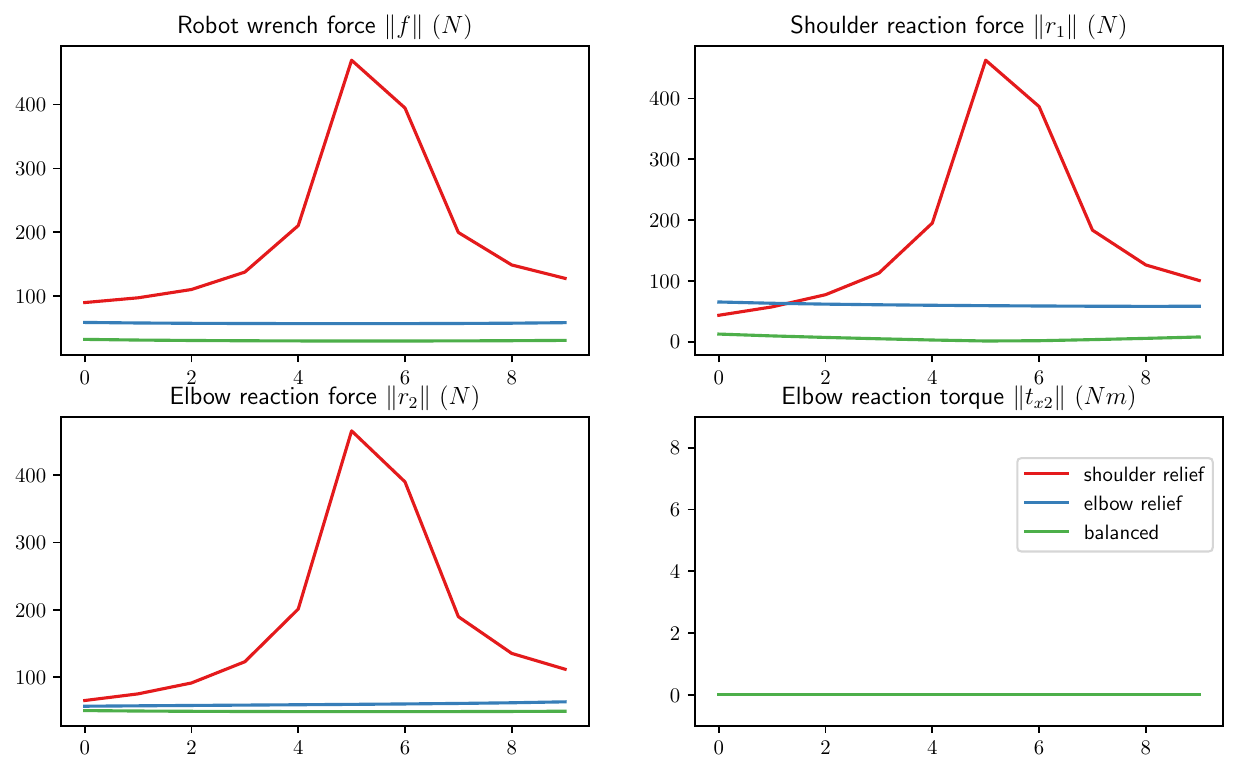}}
\caption{Forces along a trajectory by time step where an arm laid in a rest pose beside the body is raised vertically in the air. The graphs show the magnitude of the robot wrench forces, shoulder reaction forces $[r_{x1},\; r_{y1},\; r_{z1}]$, elbow reaction forces $[r_{x2},\; r_{y2},\; r_{z2}]$, and elbow reaction torques $t_{x2}$ respectively. Three different force-distribution constraints explained in Section \ref{subsec:human_force_modeling} are compared. The imbalanced force constraints result in the reaction forces beyond the acceptable values derived in Section \ref{subsec: Safety Constraints}.}
\label{fig:simulation_forces}
\end{figure}

\begin{table}[t]
\centering
\caption{Analysis of the generated trajectories} 
\label{tab:path_length_run_time}
\resizebox{0.49\textwidth}{!}{%
\footnotesize
\begin{tabular}{clccccc}
    \toprule
    \multicolumn{2}{c}{Trajectory Number} & 1 & 2 & 3 & 4 & 5 \\
    \midrule
     & Pos. Length (m) & 1.47 & 1.18 & 1.47 & 3.1 & 1.82 \\
    \textit{MP} & Ori. Length (rad) & 3.16 & 1.23 & 2.09 & 4.13 & 2.99 \\
     & Run Time (s) & 100.69 & 104.45 & 102.95 & 101.82 & 111.61 \\
    \midrule
     & Pos. Length (m) & 1.18 & 0.44 & 0.91 & 1.28 & 1.55 \\
    \textit{TO} & Ori. Length (rad) & 1.37 & 0.38 & 0.98 & 1.2 & 2.39 \\
     & Run Time (s) & 2.37 & 31 & 31.36 & 47.66 & 72.64 \\
    \midrule
    \multicolumn{2}{c}{Feasible} & \checkmark & \checkmark & \checkmark & \checkmark & \checkmark \\
    \bottomrule
    \end{tabular}
}
\end{table}

Given different sets of initial and target human-limb configurations, we first evaluate the path length, run time, and geometric feasibility after generating human-limb and robot-arm trajectories using Algorithm \ref{alg:trajectory_generation}. 
Table \ref{tab:path_length_run_time} demonstrates (1) the position as well as orientation length of the human-wrist trajectories, (2) run time, and (3) if the trajectories are geometrically feasible, i.e., if they lie within joint limits and are collision-free.
The geometric feasibility is evaluated using CoppeliaSim. 
The results show that motion planning can take a while to find a feasible path and is not guaranteed to find an optimal one given limited time. 
With the refinement stage, the quality of an initial path from motion planning can be quickly improved while still maintaining its feasibility. 

We also analyze the reaction forces/torques of the human limb and the robot wrench forces when running each trajectory. 
To demonstrate the importance of properly modeling the distribution of reaction forces on human joints, we compare the results calculated based on three different distributions as introduced in Section \ref{subsec:human_force_modeling}, i.e., shoulder relief ($r_{z1} = 0$), elbow relief ($r_{z2} = 0$), and balanced (Eq. (\ref{eq:3D force model 1})). 

Fig. \ref{fig:simulation_forces} shows the human reaction forces/torques and the robot wrench forces of a trajectory that repositions a human arm from resting beside the body to be held up vertically off the ground. 
If an imbalanced distribution is chosen, the human-joint reaction forces can be very high along the trajectory, especially when using the shoulder-relief model. 
These high values indicate that the robot arm needs to move the human limb with large wrench forces, which can cause severe damage to the human body. 
On the other hand, if we choose to balance the reaction forces on both the shoulder and the elbow (balanced), the human-joint reaction forces satisfy the safety values in Eq. (\ref{ForceLimits}). 
Note that the elbow reaction torques are very small for all three models since the arm is lifted off the ground without bending the elbow too much. 
These results demonstrate that the human-joint reaction forces predicted by our model align well with intuition. 

\subsection{Live Experiments}

\begin{figure}[t]
\vspace{2mm}
\centerline{\includegraphics[width=40mm]{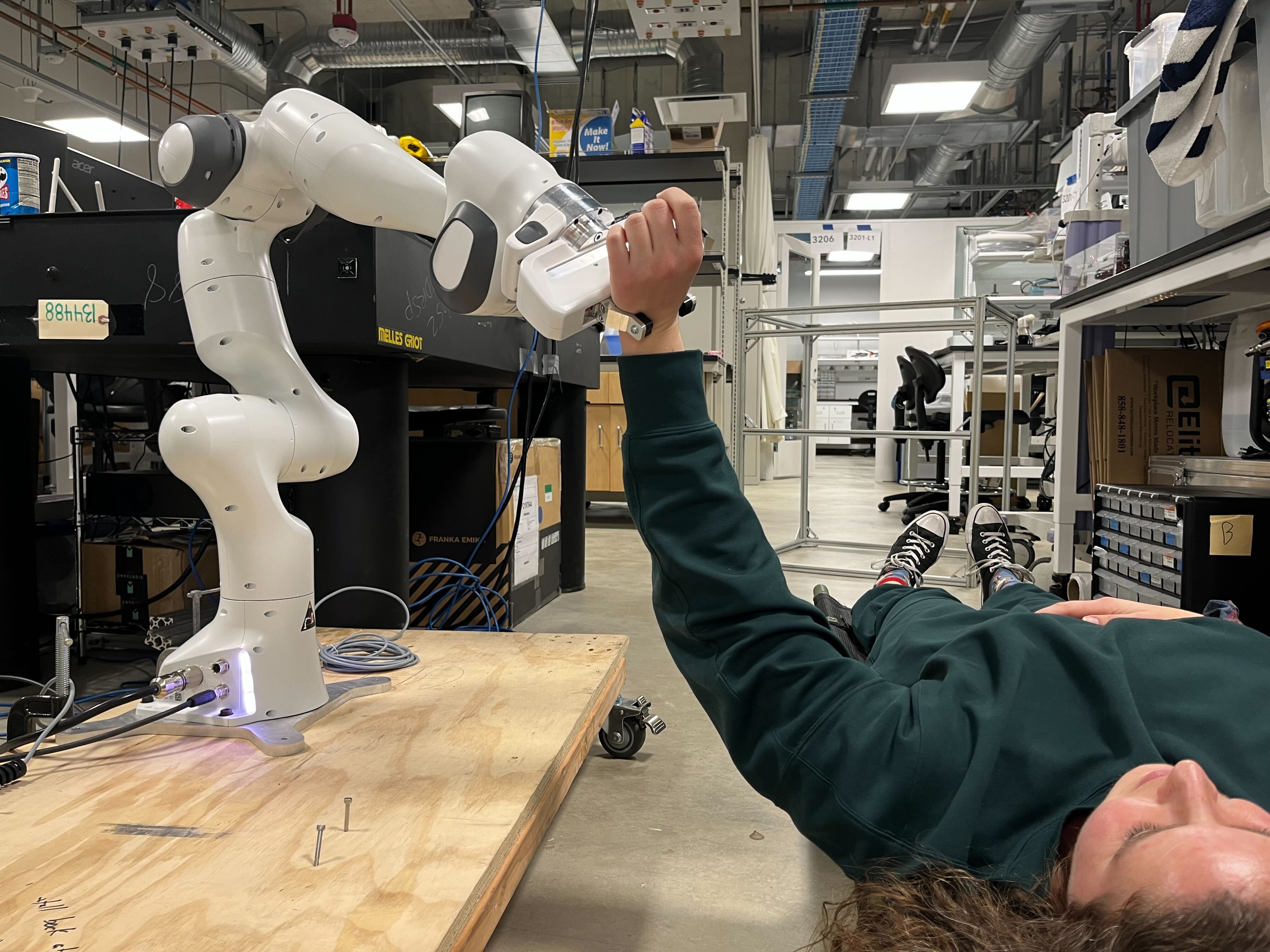} \includegraphics[width=40mm]{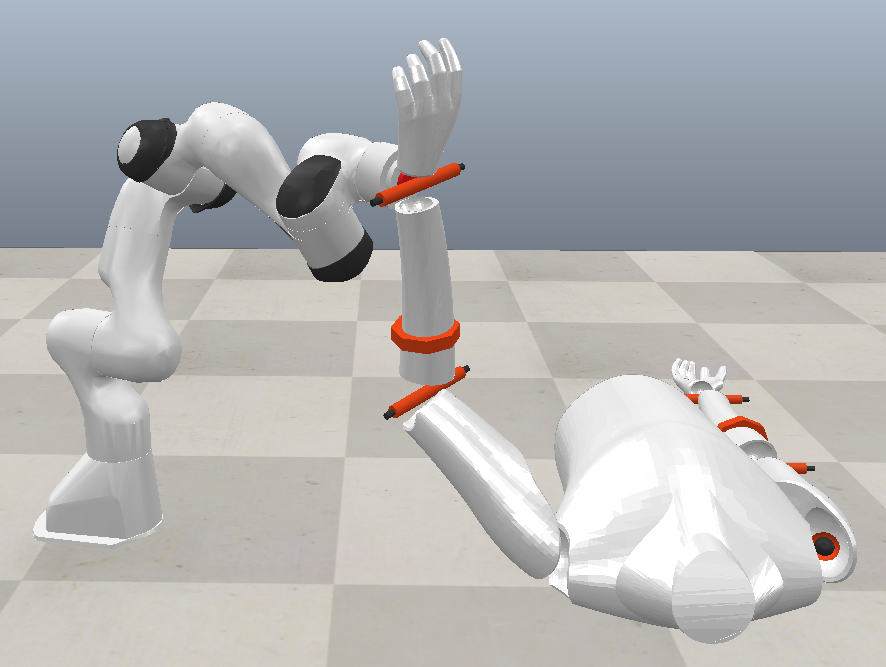}}
\caption{The images from left to right are the live human experiment including the Panda arm mounted on a mobile platform and the matching simulated trajectory shown in Coppeliasim.}
\label{Experiments}
\end{figure}

\begin{figure}[t]
\centerline{\includegraphics[width=\linewidth]{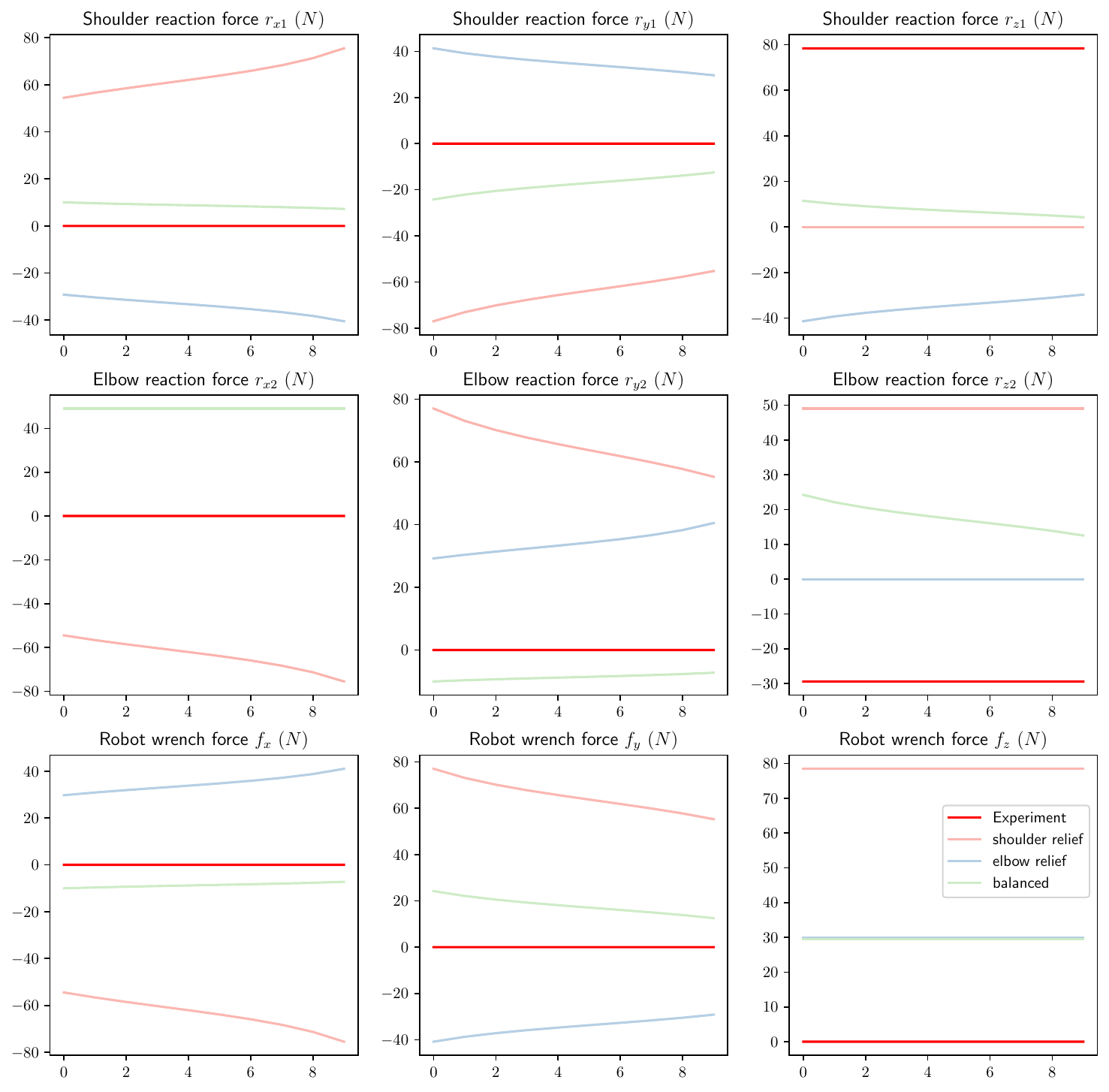}}
\caption{The forces from live experimental data. The graphs above show the shoulder-joint reaction forces $[r_{x1}, r_{y1}, r_{z1}]$, elbow-joint reaction forces $[r_{x2}, r_{y2}, r_{z2}]$, and robot wrench forces $[f_x, f_y, f_z]$ respectively with the x-axis having relative time step. Each graph contains the experimental data collected with a real human subject and the simulation data from the same trajectory using three different force-distribution constraints explained in Section \ref{subsec:human_force_modeling}. $t_{x2}$ is not graphed. Its range from the live experiment is $[-1.3, -1.5]~Nm$, and the simulated values are almost $0~Nm$.}
\label{fig:real_exp_forces}
\end{figure}

The feasible robot-arm trajectories are tested in real-world environments, where a Franka Emika Panda robot repositions a real human limb (Fig. \ref{Experiments}). 
To evaluate the robot wrench forces and the human reaction forces in live experiments, we measure the robot Jacobian matrix, $J^R(\bm{q}_i)$, and torques, $\bm{\tau}_i^R$, for each time step $i$. 
This allows us to obtain the robot wrench forces by $\textbf{w}^R_i = J^R(\bm{q}_i) \bm{\tau}_i^R$. 
Then the human reaction forces are calculated by solving Eq. (\ref{eq: short human equation}). 
Fig. \ref{fig:real_exp_forces} plots the human-joint reaction forces and the robot wrench forces from simulation and real-world data. 
Similar to the results in Fig. \ref{fig:simulation_forces}, the balanced distribution model produces the smallest predicted human-joint reaction forces compared to the shoulder-relief and elbow-relief models. 
In addition, the forces predicted by the balanced constraint are closer to the ones calculated from real-world data. 
Note that the magnitude of the robot wrench forces measured from real-world data is very small. 
The reason might be that the human arm is not completely passive, and the muscles and tendons can provide extra forces to support the arm.

\section{Discussion and Conclusions}
In this paper, we explored a critical aspect of search and rescue operations of unconscious casualties carried out by mobile manipulators: reconfiguring limbs of a human body under biomechanical safety constraints. We provide the first formal representation of the overall robot motion planning problem and present a solution to solve the geometric and biomechanical constraints that arise. 

Several limitations arise from this current initial approach.
For problem formulation, the breakup of the problem solving for intermediate wrench forces on the robot guarantees sufficiently low joint loading. 
However, the robot trajectory is still defined using kinematic constraints. 
These constraints do not limit the wrench forces outputted from the robot, and the position controller within the robot defines its own force limits. 
Our solution solves trajectories as a quasi-static problem and loosely goes between statics and geometric constraints/losses. 
A more ideal solution integrates dynamics rather than quasi-static assumptions into path planning and optimization. 
The choice to constraint joint force components, although reasonable, limits the possible solutions for acceptable wrench forces of optimizing the robotic trajectory and rejects some paths that would otherwise have been acceptable. 
Furthermore, the tendon anatomy, specifically its effect in applying spring forces that provide resistive forces within the joint range of motion, was not accounted for and may have impacted force distribution in the live demo results compared to our simulated model. Finally, our algorithm can be sped up by porting everything from Python to C++.

Future work will involve addressing other components of the larger casualty repositioning problem, including identifying a discrete sequence of key human poses that move from one pose to another (i.e., a task planning problem), determining biomechanically safe grasp candidates on limbs with consideration of injuries, and accounting for partial or noisy observations in real camera settings.
\clearpage
\bibliographystyle{ieeetr}
\bibliography{refs}

\end{document}